\documentclass[sigconf,arxiv]{acmart}
\AtBeginDocument{%
  }

\setcopyright{acmlicensed}
\copyrightyear{2018}
\acmYear{2018}
\acmDOI{XXXXXXX.XXXXXXX}
\acmConference[Conference acronym 'XX]{Make sure to enter the correct
  conference title from your rights confirmation email}{June 03--05,
  2018}{Woodstock, NY}
\acmISBN{978-1-4503-XXXX-X/2018/06}
\usepackage{enumitem}
\usepackage{multirow}
\usepackage{balance}
\usepackage[normalem]{ulem}
\usepackage{graphicx}
\usepackage{caption}
\usepackage{subcaption}
\usepackage{array}
\useunder{\uline}{\ul}{}

\begin{document}


\title{GIST: Cross-Domain Click-Through Rate Prediction via Guided Content-Behavior Distillation
}



\author{Wei Xu}
\authornotemark[1]
\affiliation{%
  \institution{Xiaohongshu Inc.}
  \city{Shanghai}
  \country{China}}
\email{weixu12231227@gmail.com
}

\author{Haoran Li}
\authornote{Both authors contributed equally to this research.}
\affiliation{%
  \institution{Xiaohongshu Inc.}
  \city{Shanghai}
  \country{China}}
\email{haoran.li.cs@gmail.com}

\author{Baoyuan Ou}
\affiliation{%
  \institution{Xiaohongshu Inc.}
  \city{Shanghai}
  \country{China}}
\email{oubaoyuan@xiaohongshu.com}

\author{Lai Xu}
\affiliation{%
  \institution{Xiaohongshu Inc.}
  \city{Shanghai}
  \country{China}}
\email{xulai@xiaohongshu.com}

\author{Yingjie Qin}
\authornote{Corresponding author.}
\affiliation{%
  \institution{Xiaohongshu Inc.}
  \city{Shanghai}
  \country{China}}
\email{yingjieqin@xiaohongshu.com}

\author{Ruilong Su}
\affiliation{%
  \institution{Xiaohongshu Inc.}
  \city{Shanghai}
  \country{China}}
\email{suruilong@xiaohongshu.com}

\author{Ruiwen Xu}
\affiliation{%
  \institution{Xiaohongshu Inc.}
  \city{Shanghai}
  \country{China}}
\email{rig@xiaohongshu.com}









\renewcommand{\shortauthors}{Wei Xu et al.}

\begin{abstract}

Cross-domain Click-Through Rate prediction aims to tackle the data sparsity and the cold start problems in online advertising systems by transferring knowledge from source domains to a target domain. Most existing methods rely on overlapping users to facilitate this transfer, often focusing on joint training or pre-training with fine-tuning approach to connect the source and target domains. However, in real-world industrial settings, joint training struggles to learn optimal representations with different distributions, and pre-training with fine-tuning is not well-suited for continuously integrating new data. To address these issues, we propose GIST, a cross-domain lifelong sequence model that decouples the training processes of the source and target domains. Unlike previous methods that search lifelong sequences in the source domains using only content or behavior signals or their simple combinations, we innovatively introduce a Content-Behavior Joint Training Module (CBJT), which aligns content-behavior distributions and combines them with guided information to facilitate a more stable representation. Furthermore, we develop an Asymmetric Similarity Integration strategy (ASI) to augment knowledge transfer through similarity computation. Extensive experiments demonstrate the
effectiveness of GIST, surpassing SOTA methods on offline
evaluations and an online A/B test. Deployed on the Xiaohongshu (RedNote) platform, GIST effectively enhances online ads system performance at scale, serving hundreds of millions of daily active users.

\end{abstract}

\ccsdesc[500]{Information systems~Recommender systems}

\keywords{Cross-domain recommendation, CTR Prediction, Multi-modal Representations}



\maketitle

\section{Introduction}


The online advertising industry has been gaining massive attention due to its substantial market value and broad applications~\cite{zhou2018deep, li2021attentive, ouyang2021learning, liu2024at4ctr}. One of the
most fundamental techniques in ads recommendation is accurate
Click-Through Rate (CTR) prediction. For example, CTR prediction can predict user engagement in specific content in various advertising formats, including sponsored search ads, display ads, and short video ads.

\begin{figure}[t!]
  \centering
  \includegraphics[width=\linewidth]{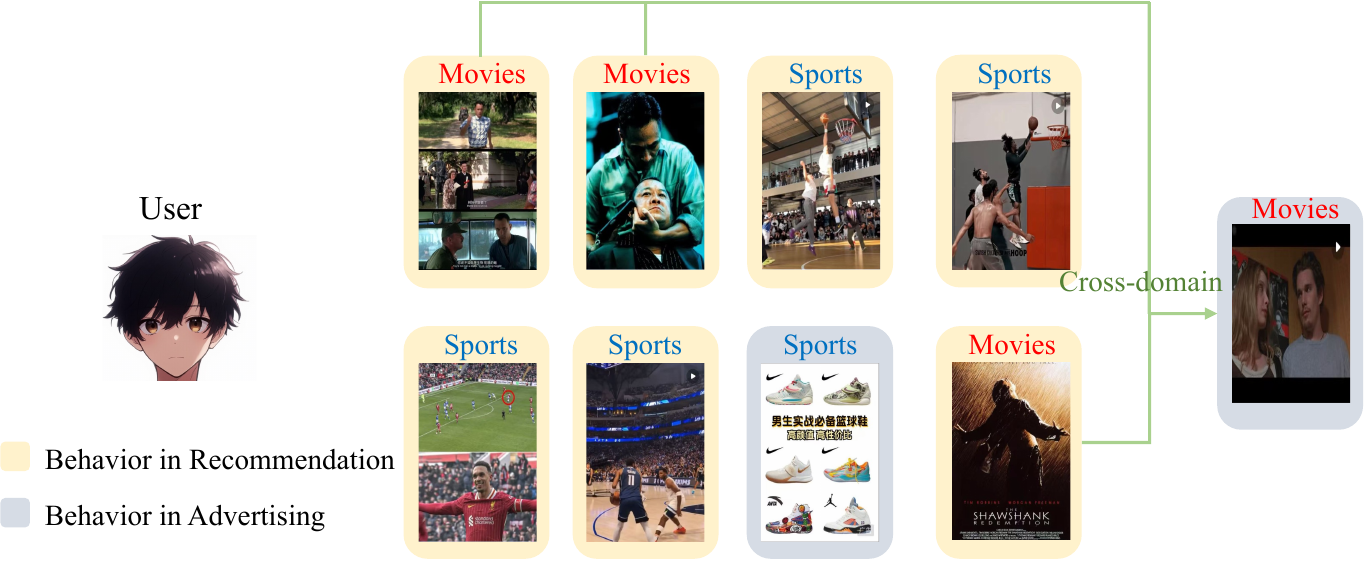}
  \caption{Illustration of cross-domain recommendation. Here, knowledge in the recommendation domain is transferred to the advertising domain.}
  \label{fig:intro}
\end{figure}

Recently, deep neural networks (DNNs) have significantly advanced the accuracy of CTR predictions~\cite{zhang2019deep, guo2017deepfm, wang2017deep, wang2021dcn}. They achieve this by learning representations of user interests from behavioral sequences, especially those related to target items~\cite{zhou2018deep, pi2020search}. However, this task has become increasingly challenging due to the data sparsity and the cold-start problems in advertising systems~\cite{ouyang2021learning, liu2024at4ctr, pi2020search}. As shown in Figure~\ref{fig:intro}, in the industrial scenario of Xiaohongshu (RedNote), the behavior sequence of a user includes
numerous items related to sports and movies; however, many of
these behaviors originate from the recommendation domain, highlighting a stark contrast with the lack of data in the advertising domain. 



To address these challenges, existing approaches have introduced cross-domain CTR prediction, which addresses the data sparsity problem by leveraging knowledge from information-rich source domains~\cite{zhu2022personalized, zhao2023cross, cao2022disencdr, zhao2023cross}. As illustrated in Figure~\ref{fig:intro}, we can transfer knowledge from the recommendation domain to the advertising domain. There are currently two common paradigms~\cite{ouyang2020minet, chen2024survey}. The first approach involves jointly training samples from both the source and target domains, utilizing multi-task learning to develop a shared entity representation, or employing contrastive learning~\cite{chen2020simple,chen2021exploring} to align representations of cross-domain entities~\cite{li2020ddtcdr,ma2018modeling}. The second approach adopts a two-stage method, where samples from the source domain are used for pre-training, followed by fine-tuning on the target domain~\cite{chen2021user, liu2024mcrpl}. However, both of these two paradigms face significant challenges. Specifically, the joint training approach often struggles to learn optimal joint representations with different distributions and unbalanced domain samples\cite{yu2020gradient}. Meanwhile, the pre-training with fine-tuning approach faces the challenge of integrating continuously generated interaction data in modern industrial systems.

To address these issues, in this paper, we introduce cross-domain lifelong behavior sequences and design a decoupled training paradigm, where items from the source domain learn embeddings independently in the target domain. Compared to the joint training and pre-training with fine-tuning approaches, the decoupled training paradigm offers greater flexibility and stability of transfer, e.g., the learning process of the target model is less disturbed by the constantly updating data in the source domain. Additionally, under this decoupled training paradigm, we expect to answer the following question: \textit{how to effectively and efficiently transfer and model lifelong behavior sequences}?

The conventional approach~\cite{pi2020search} for managing user behavior sequences divides the modeling process into two units: the General Search Unit (GSU) and the Exact Search Unit (ESU). The GSU aims to search a small subset of items most relevant to the target item from the user behavior sequences, thereby reducing computational complexity for subsequent  models. The ESU then performs fine-grained modeling on the filtered items from GSU to  extract more precise user interest representations. Currently, the dominant GSU method in the industry is soft search, which retrieves items by computing similarities between vector representations. While many implementations reuse item embeddings learned during model training—demonstrating strong retrieval performance for well-trained items—these embeddings often underperform for long-tail items or those in cold-start states, limiting overall GSU efficacy. To address this limitation, some methods~\cite{yuan2023go} have explored the potential of multi-modal vectors, which show notable advantages for items with sparse interaction signals. This has made multi-modal retrieval a standard practice for cold-start scenarios. However, a key trade-off persists: multi-modal vectors, while robust for items with limited interaction data, often lack sufficient collaborative signals to fully exploit the retrieval potential of items with abundant user interactions. This underscores a critical need for the adaptive integration of multi-modal (content) and collaborative (behavior) signals to balance retrieval performance across cold-start and interaction-rich items.

To address these challenges, we propose GIST, a \textbf{G}u\textbf{i}ded Content-Behavior Di\textbf{st}illation framework for cross-domain CTR prediction. GIST decouples source and target domain training while synergistically integrating multi-modal content signals and user behavioral signals. Specifically, we first design a Content-Behavior Joint Training Module (CBJT) to learn robust content-behavior joint representations, achieving balanced retrieval performance for both popular and long-tail items. It is worth noting that we select high-quality item-to-item (i2i) pairs from the source domain ESU as the training data (guidance) for CBJT, aiming to distill fine-grained selection preferences from the source domain. Then, in the target domain, we employ the joint representations from CBJT as the foundation of the GSU, which is used to search lifelong sequences by cosine similarity. For the searched items, we adopt independent learning techniques to acquire representations specific to the target domain, rather than fine-tuning representations from the source
domain or employing contrastive learning to align with them. Furthermore, we propose an Asymmetric Similarity Integration strategy (ASI) that leverages the joint representation to compute cosine similarity scores and similarity distributions between target items and the selected historical interactions. These computed metrics are then utilized as input features to predict CTR in the target domain.



%






In summary, the main contributions of this paper are as follows:
\begin{itemize}[leftmargin=0.5cm]
    \item We propose a novel GIST framework to address the data sparsity problem in the advertising domain by transferring knowledge from the lifelong behavior sequences in the recommendation domain. Moreover, we propose to decouple source and target domain training, facilitating transfer flexibility and stability.
    \item We present CBJT, a Content-Behavior Joint Training module that enables high-quality knowledge transfer through two types of alignments: (1) alignment between the content and behavioral signals through behavior-based encoder; (2) alignment between selection preferences of the GSU and ESU through high-quality i2i pairs from the ESU in the source domain (recommendation) ranking model.
    \item We validate the effectiveness of our method through extensive offline experiments and ablation studies. Moreover, online experiments show that our approach significantly boosts advertising revenue on a real-world industrial platform with hundreds of millions of daily active users.
\end{itemize}

\begin{figure*}[t]
  \centering
  \includegraphics[height=0.4\textwidth,width=\textwidth]{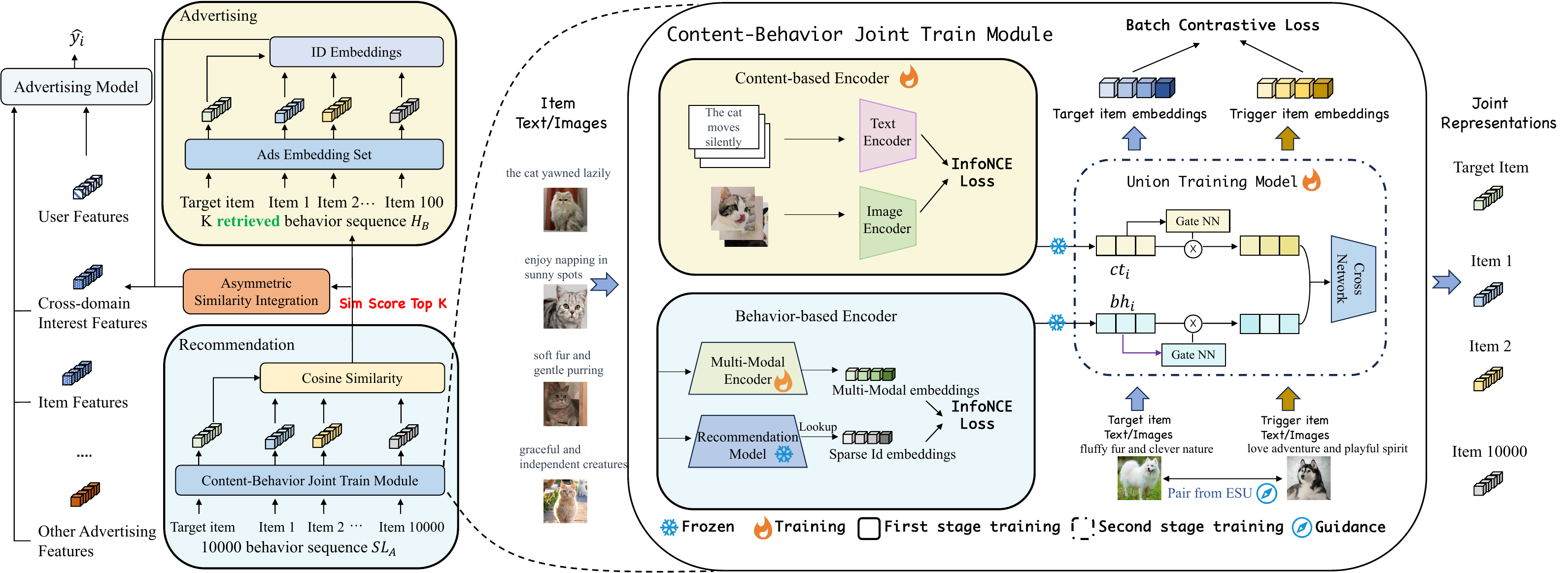}
  \caption{Overview of GIST, a Guided
  Content-Behavior Distillation framework designed to facilitate knowledge transfer in cross-domain CTR prediction. There are two major components in the model: (a) The Content-Behavior Joint Training module (CBJT) combines content and behavioral signals to produce a stable joint representation, which is used to search lifelong behavior sequence from the source domain to the target domain. (b) The Asymmetric
  Similarity Integration strategy (ASI) integrates similarity information to augment knowledge transfer, which is detailed in Figure~\ref{fig:simi}.
  }
  

  %



  \label{fig:arch}
\end{figure*}

\section{Preliminaries} 

In this section, we introduce the notations used in this paper and formally define the problem setup.

\textbf{Contrastive Learning.} The contrastive learning paradigm has been proven to be effective in aligning different modalities (distributions) \cite{yang2021taco, yang2022vision}. Take CLIP~\cite{radford2021learning} as an example, for a batch of $N$ paired images and text, the image and text features $\{ x_I, x_T\}$ are obtained using corresponding encoders. Then, the image and text features are used to compute the InfoNCE loss~\cite{oord2018representation}, where the paired images and text form the positive pairs and the unpaired ones are
treated as negative samples:


\begin{equation}
    L_I=-\sum_{i=1}^{N} \frac{exp(sim(x_I^i,x_T^i)/ \tau )}{\sum_{k=1}^{N} exp(sim(x_I^i,x_T^k)/ \tau)}
    \label{eq:con}
\end{equation}


where $(x_I^i, x_T^i)$ denotes the features of the $i^{th}$ image-text pair, $sim(\cdot, \cdot)$ denotes the similarity measurements, and $\tau$ is a learnable temperature parameter. A symmetrical loss $L_T$ is also computed, and the total loss is $L = (L_I + L_T )/2$.


\textbf{Cross-Domain Recommendation (CDR).} CDR aims to leverage the knowledge of user behavior sequences in the source domain to aid CTR prediction in the target domain, which shares overlapping user sets with the source domain but has limited behavioral data. For clarity in the subsequent sections, we define the following concepts:
\begin{itemize}[leftmargin=0.5cm]
    \item The $i$-th user is denoted as $u_{i}$, and the corresponding target items in the source and the target domain are denoted as $v_{i}^s$ and $v_{i}$, respectively.
    \item The lifelong behavior sequences in the source domain: $SL_{A}=\{sl_{1}, sl_{2},...,sl_{L} \}$.
    \item Short behavior sequences in the source domain that need to be computed by the target attention module after searching: $S_{A}=\{s_{1},s_{2},...,s_{R}\}$.
    \item Short behavior sequences in the target domain that need to be computed by the target attention module after searching: $H_{B}=\{h_{1},h_{2},...,h_{R}\}$.
\end{itemize}

\textbf{Problem Definition.} For a given user $u_{i}$ and the target item $v_{i}$ in the target domain, our goal is to predict the CTR of user $u_{i}$ on item $v_{i}$:
\begin{equation}
    p_i=P\left(y_i=1 \mid u_i, v_i, o_{i}, H_{B}, S_{A}, SL_{A}; \theta\right)
\end{equation}
where $y_{i}$ represents the user's actual feedback, $o_{i}$ denotes the user's attributes other than the sequence representation (such as age and gender), and $\theta$ denotes the model parameters. The objective function can be formulated as:
\begin{equation}
\mathcal{J}=\text { minimize }\left\{\frac{1}{N} \sum_{i=1}^{N} \operatorname{CrossEntropy}\left(y_i,p_{i}\right)\right\}    
\end{equation}
where $N$ denotes the number of samples in the training set.

\section{Methodology} 

In this section, as illustrated in Figure~\ref{fig:arch}, we will introduce GIST, a \textbf{g}u\textbf{i}ded
content-behavior di\textbf{st}illation framework designed to decouple the representation learning in the source and target domains in cross-domain recommendation. Specifically, GIST consists of two complementary modules. First, we propose a Content-Behavior Joint Training Module (CBJT) to learn stable content-behavior representations. To elaborate, CBJT implements two types of alignments: (1) alignment between the content and behavioral signals through a behavior-based encoder; (2) alignment
between the selection preferences of the General Search Unit (GSU) and Exact Search Unit (ESU) through high-quality i2i pairs from the ranking ESU process. Subsequently, we use the content-behavior representations from CBJT to search top-$K$ items with the highest similarity scores to the target item in the target domain. It is worth noting that we do not directly forward the item embeddings. Instead, we only forward the item IDs and their similarity scores, aiming to \textbf{decouple} the training of the source and target domains. Second, we design an Asymmetric Similarity Integration strategy (ASI) to augment knowledge transfer by integrating similarity scores and distributions. Next, we will detail these key modules in detail as follows:

\subsection{Content-Behavior Joint Training Module}


Behavior sequences capture users' interactions during their engagement with a platform. When focusing on a target item, only a subset of these sequences may significantly influence the user's CTR. For lifelong sequences that can extend to tens of thousands ($10^4$) in length, it's essential to pinpoint the few hundred ($10^2$) items that are most pertinent to the target item. This selective process not only optimizes the model's capacity but also enhances computational efficiency. Conventional sequence modeling systems generally divide interest extraction into two stages: the GSU and the ESU. The GSU is responsible for searching the sequence and identifying the items most relevant to the target item, while the ESU applies sophisticated attention mechanisms to further hone the user's interest preferences from the GSU-selected items. The efficacy of the GSU significantly relies on the quality of item embeddings. Prior methods fall into two categories: one uses multi-modal representations without integrating user behavior information, relying solely on content-based data; the other reuses item embeddings derived during model training, which often under-perform with items of low interactions or in cold start scenarios.


To address these challenges, we introduce a Content-Behavior Joint Training Module (CBJT), as illustrated in the right part of Figure~\ref{fig:arch}. This module combines behavioral signals and content signals for joint training, creating high-quality item embeddings. By integrating text and image data as inputs, the module employs a content-based encoder and a behavior-based encoder to generate multi-modal embeddings from both content and behavior perspectives. The union training model distills item-to-item (i2i) pairs, synthesizing outputs from both encoders to produce a joint representation. This method not only captures user interest from behavior data but also infuses content information, yielding richer and more precise item embeddings, and thereby boosting the overall model performance and predictive accuracy.


\subsubsection{\textbf{Content-based Encoder}}

With recent advancements in multi-modal learning~\cite{radford2021learning, li2022blip, li2023blip, li2024lite, li2025hope}, we propose to capture content-level information through a widely adopted multi-modal pre-training framework, which consists of a text encoder and an image encoder. These encoders encode the input text and image into respective feature vectors. During training, each image is paired with its corresponding text description to form a positive sample pair, while random pairing of images and texts within the same batch serves as negative sample pairs. We utilize a contrastive loss function, such as the InfoNCE loss ~\cite{oord2018representation}, to optimize the model by maximizing the similarity between positive sample pairs and minimizing the similarity between negative ones as in Equation~\ref{eq:con}. This approach enables the model to learn unified multi-modal representation $ct_{i}$ for the content data: 
\begin{equation}
    ct_i = F_{C} (v_i)
\end{equation}
where $F_{C}$ denotes the content-based encoder, $v_i$ denotes the image and text data of the item (here we use the target item $v_i$ for simplicity).

\subsubsection{\textbf{Behavior-based Encoder}}

Aside from learning representations from content data, we also propose to capture the dynamics of user behavior sequences. Specifically, when reusing item embeddings learned during model training for the GSU, we've observed that highly interacted items, which undergo extensive training, develop embeddings rich in behavioral information. These embeddings also exhibit strong retrieval capabilities, facilitating efficient searching. Conversely, long-tail items suffer from insufficient training, leading to lower-quality embeddings that adversely affect the GSU's overall performance. Motivated by these findings, we aim to make full use of the item ID embeddings obtained from the source domain. Specifically, we first select items with interaction counts above a certain threshold as reliable signals. Then, we encode the content data using a multi-modal encoder, which is a separate copy of the content-based encoder. Subsequently, we lookup the item ID embeddings from the source domain. During training, each multi-modal embedding $bh_i$ is paired with its corresponding sparse ID embedding $id_{i}$ to form a positive
sample pair, while random pairing within the
same batch serves as negative sample pairs. We utilize a contrastive
loss function, i.e., the InfoNCE loss [22], to optimize the behavior model:
\begin{align}
    L_C & =  -\sum_{i=1}^{N} \frac{exp(sim(bh_i,id_i)/ \tau )}{\sum_{k=1}^{N} exp(sim(bh_i,id_k)/ \tau)}
    \\[1em]
    L & = \frac{L_C + L_B}{2}
\end{align}
where $L_B$ denotes the symmetrical loss of $L_C$. This process aligns the behavior and content distributions, ultimately producing the behavior-based representation $bh_{i}$ in the inference stage:
\begin{equation}
    bh_i = F_{B} (v_i)
\end{equation}
where $F_{B}$ denotes the behavior-based encoder, $v_i$ denotes the image and text data of the item (here we use the target item $v_i$ for simplicity).






    

\subsubsection{\textbf{Union Training Model}}
After obtaining representations from content data and user behaviors, we propose a Union Training Model to fuse Content-Behavior representations. 

The two encoders mentioned above produce content-based and behavior-based representations, respectively. To effectively fuse these two sets of representations, we propose a unified representation learning approach. Specifically, we find that when estimating relevance, ESU cannot accurately rank all interacting items. Previous work~\cite{feng2024long} issues a similar viewpoint. However, it demonstrates strong capability in identifying the most relevant item. Therefore, we choose to distill ESU pairs consisting of the target and the most relevant item to enhance the fusion process.

The ESU typically serves as the target attention module. We compute attention scores between each target item $v_{i}^s$ and the user's historical interaction items in source domain $S_{A}=\{s_{j}\}_{j=1}^{R}$ and choose the most relevant item:
\begin{equation}
\begin{array}{cl}
    a_{j}=\operatorname{Attention\left(v_{i}^s,s_{j} \right)}, \\
    k=\operatorname{argmax}(a_{j}),\forall j \in \{ 1,2,\dots, R\}
\end{array}
\label{equ:ak}
\end{equation}

Then we organize this data and form pairs $(v_{i}^s, s_{k})$ for items meeting the condition $a_{k}\geq \theta$. These qualifying pairs act as confidence data for the fusion training. 

The joint training model comprises two components. The first component employs a gating mechanism~\cite{swietojanski2016learning} to judge feature importance:
\begin{equation}
    \begin{array}{cl}
    f_{c t}=\operatorname{MLP}_{ct}\left(ct_i\right), & c \hat{t}_i=2 \cdot \operatorname{sigmoid}\left(f_{ct}\right) \cdot ct_i \\
    f_{b h}=\operatorname{MLP}_{bh}\left(bh_i\right), & b \hat{h}_i=2 \cdot \operatorname{sigmoid}\left(f_{bh}\right) \cdot bh_i
    \end{array}
\end{equation}

The second component combines the two representations into one using a cross network. By employing a commonly used tensor fusion method, we chose the tensor outer product approach~\cite{zadeh2017tensor} to enhance the model's expressive capability:

\begin{equation}
    u_{i}=\operatorname{MLP}_{cs}(\hat{ct_{i}} \otimes \hat{bh_{i}})
\end{equation}
where $\otimes$ indicates the outer product between vectors and $u_{i}$ is the joint representation of the item.

Finally, we align the previously distilled ESU pairs and optimize the model using an in-batch contrastive loss function. This approach enables the model to more effectively integrate content and behavior representations.

\begin{figure}[t]
  \centering
  \includegraphics[width=0.85\linewidth]{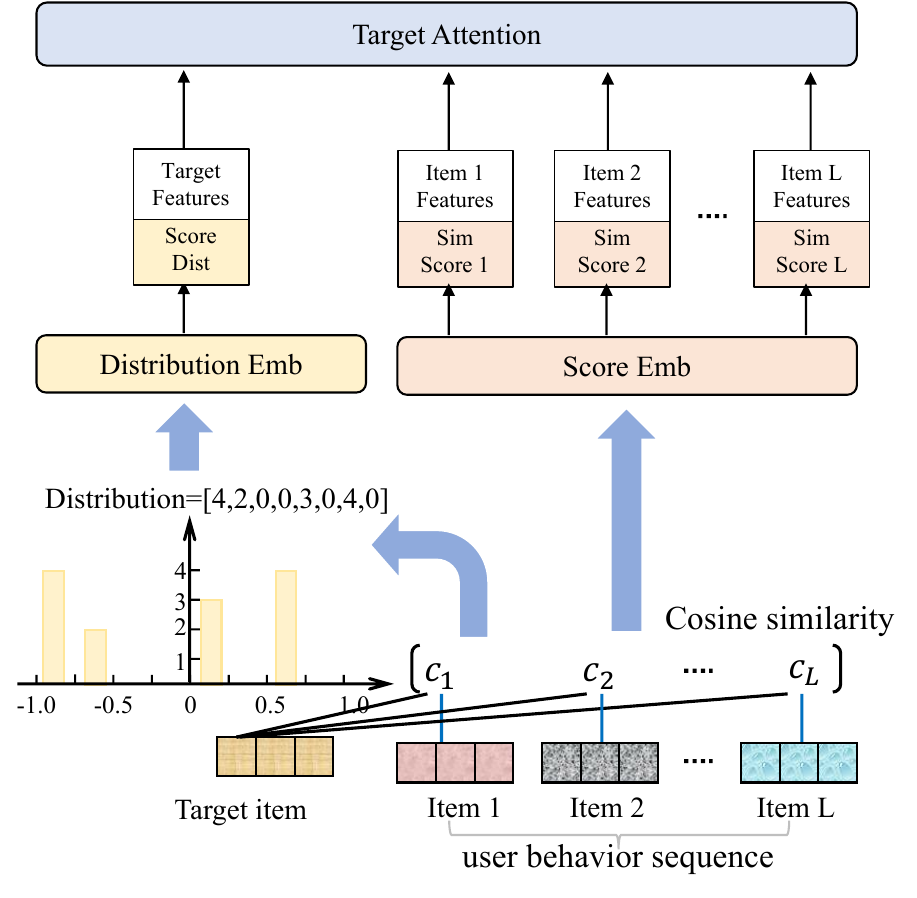}
  \caption{Structure of the Asymmetric Similarity Integration strategy (ASI). The similarity score of each item, denoted as $c_i$, along with the similarity distribution of the target item, is embedded and concatenated as features within the ESU.}
  \label{fig:simi}
\end{figure}

\subsection{Integration with Advertising Models}

To effectively integrate items closely associated with the target item into the advertising model, a strategic approach is essential. Traditionally, methods like contrastive learning have been employed to align embeddings between the recommendation and advertising domains. However, such methods can disrupt the natural data distribution within each domain. Therefore, we've opted to decouple the representation relationships in these domains, allowing the advertising domain to independently learn item embeddings. This strategy enhances the capturing of cross-domain user interests.

In addition to its role as a lifelong sequence searcher, the joint representation should also be integrated into the advertising model using other methods. One straightforward approach is to directly incorporate it into an ID-based model, concatenating the joint representation with the item's ID embedding. Furthermore, previous research~\cite{sheng2024enhancing} suggests that it often underperforms because the parameters associated with the joint representation are not adequately learned during joint training with ID embeddings. Instead, simplifying the use of joint embeddings has been shown to improve performance.

Therefore, we focus on using similarity strength as the core information of the representation and propose an asymmetric integration approach. As depicted in Figure \ref{fig:simi}, we first calculate the cosine similarity between each target item $v_{i}$ and the user's historically interacted items in target domain $H_{B}=\{h_{j}\}_{j=1}^{R}$:
\begin{equation}
c_{j}=u_{i} \cdot u_{j} ,\forall j \in \{1,\dots,R\}
\end{equation}
where $u_{i}$ is the joint representation of the item. We then incorporate similarity information for both the target item and the user's historical interactions to streamline the application of joint embeddings.

\textbf{Similarity Score Embedding:} We partition  $c_{j} \in [-1.0, 1.0]$ into $M_{1}$ intervals, converting $c_{j}$ into an ID form through discretization. Representations are learned using the embedding set and are concatenated into the users' historically interacted items' representation within the target attention module.

\textbf{Similarity Score Distribution:} We divide $c_{j} \in [-1.0, 1.0]$ into $M_{2}$ intervals. For the target item $v_{i}$, we tally the number of user interaction items falling within each interval, creating an $M_{2}$-dimensional vector whereby each dimension corresponds to the count of similarity scores within that interval. This process effectively transforms a set of high-dimensional multi-modal representations into an $M_{2}$-dimensional vector encapsulating the similarity distribution between the target item and the user's historical interactions. We learn the representation of this vector as part of an embedding set, integrating it into the representation of the target item within the target attention module.

In summary, this approach introduces recommendation domain information into the advertising domain via a lifelong sequence and effectively integrates cross-domain information through sequence searching and similarity computation, thereby enhancing the advertising model's performance.

\begin{figure}[t]
  \centering
  \includegraphics[width=\linewidth]{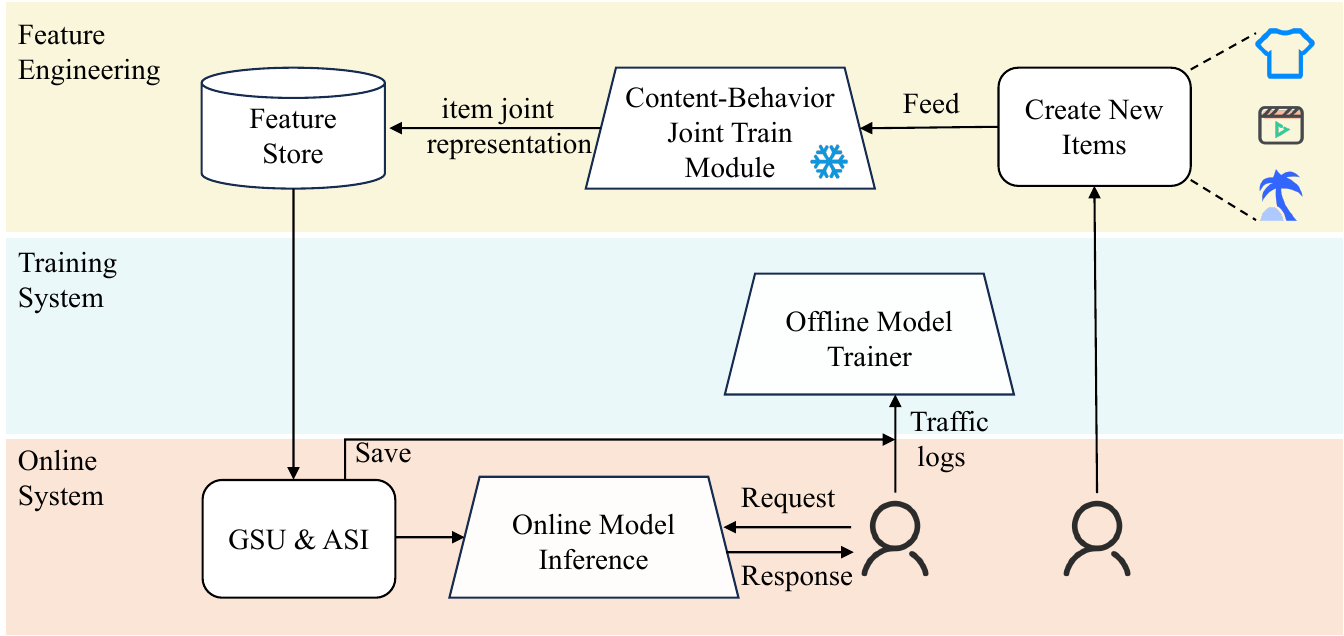}
  \caption{An overview of GIST workflow in the online system.}
  \label{fig:online}
\end{figure}

\section{Industrial Design for Online Deployment}
In industrial settings, new items are continually being developed. To ensure accurate predictions for these new items, it is essential for CTR models to acquire real-time joint representations of them. An illustrative overview of our online system is shown in Figure~\ref{fig:online}. Similar to traditional multi-modal models, when new items are introduced, the system automatically requests the pre-trained Content-Behavior Joint Train Module to compute joint representations of these items. These representations are then sent to the index tables of the feature storage center, enabling downstream online systems to retrieve joint representations from the storage system, thus facilitating real-time online prediction capabilities. It is important to note that, like multi-modal modules, the CBJT is pre-trained and asynchronously stored, which does not increase inference delay; it simply replaces previous multi-modal representations with joint representations. At the end of each inference, the results from the GSU, specifically the top-k historical interaction items corresponding to the target item, are directly stored as features in the traffic logs. As a result, there are no additional computational costs incurred during training. The only computational expense added by GIST is within the ASI module, where a minimal number of parameters are added for similarity information recognition, resulting in a total delay increase of only 0.1\%.

\section{Experiments}

In this section, we conduct offline and online experiments on large-scale industrial dataset to answer the following research questions:
\begin{itemize}[noitemsep, topsep=0pt,left=0cm]
    \item \textbf{RQ1:} How does GIST perform compared to SOTA methods?
    \item  \textbf{RQ2:} Can the proposed modules (e.g., content-behavior joint training module) effectively improve performance?
    \item \textbf{RQ3:} How does GIST affect the performance of items with different levels of popularity?
    \item  \textbf{RQ4:} How effective is GIST in real-world online advertising systems?
\end{itemize}

\subsection{Experimental Setup}
\subsubsection{\textbf{Dataset}}
The dataset was collected from traffic logs of the Xiaohongshu (RedNote) platform and includes user behavior sequences related to both advertisements and recommendations. For each user, we gathered their behavior sequences in the advertising scenario (with a maximum length of 200) as well as their lifelong behavior sequences in the recommendation scenario (with a maximum length of 10,000). Each sample is labeled by the user's click actions on ad items. The dataset comprises 1.6 billion records from 100 million users, collected over a span of 7 days. We temporally partitioned the dataset, using data from the first 6 days for training and the data from the 7th day as the test set.

\subsubsection{\textbf{Baselines}}
To evaluate our proposed GIST, we select an array of state-of-the-art (SOTA) methods for comparison. Our initial baseline was established without introducing cross-domain sequences. Below is an overview of the competing methods:
\begin{itemize}
    \item \textbf{DIN}~\cite{zhou2018deep}: Do not introduce cross-domain behavior sequences; only model the advertising scenario.
    \item \textbf{SIM Hard}~\cite{pi2020search}: 
    Introduce cross-domain lifelong behavior sequences. During the GSU phase, use a hard search method (e.g., search items of the same category) for selection.
    \item \textbf{SIM Soft (pooling)}~\cite{pi2020search}: During the GSU phase, use a soft search method with multi-modal embeddings for selection and perform average pooling operations on searched items to embed them into the model.
    \item \textbf{SIM Soft (attention)}~\cite{pi2020search}:Base on SIM Soft (pooling), introduce ESU and use target attention to model the items searched by GSU.
    \item \textbf{GIST}: 
    During the GSU phase, use joint representations for searching; during the ESU phase, introduce similarity information as sideinfo.

\end{itemize}

\subsubsection{\textbf{Metrics}}



Following previous work\cite{deng2024end}, we use AUC as the primary performance indicator for offline evaluation. Additionally, to assess the retrieval performance of the joint representations, we index the joint representations of all items, using ESU pairs as the ground truth. We evaluate the performance using Recall@K at various K levels, which measures the proportion of target items retrieved among the top-$K$ most relevant items. For the online A/B test, we consider three key metrics for the ads system: the CTR, the Cost Per Mille (CPM), and the Income.


\begin{table}[]
\centering
\caption{Offline performance.}
\label{tbl:offline}
\centering
\begin{tabular}{c | ccc}
\toprule[2pt]
Method              & AUC & AUC gain \\ \midrule 
DIN                 & 0.7666 & -         \\
SIM Hard            & 0.7687 & 0.273\%        \\
SIM Soft (pooling)   & 0.7695 & 0.377\%      \\
SIM Soft (attention) & 0.7701 & 0.454\%      \\
GIST     & 0.7720 & 0.699\%      \\
\bottomrule[2pt]
\end{tabular}
\end{table}

\subsubsection{\textbf{Backbones}}
We utilize the BLIP-2~\cite{li2023blip} framework for our multi-modal training, incorporating both Image-Text Contrastive Learning (ITC) and Image-Text Matching (ITM) techniques. We utilize the ViT-B-16 as the vision encoder, and choose the RoBERTa-wwm-ext-base-chinese as the text encoder. Both have been proven to be powerful single-modal models in practice.

\subsubsection{\textbf{Parameter Settings}}
The values of the hyperparameters that we select are as follows: for the behavior-based encoder, we select sparse ID embeddings with interaction counts greater than 200,000 as the alignment target. For the union training model, we extract pairs with attention scores greater than 0.4 from the ESU as training data. For the cross-domain GSU, we select the top 100 behaviors.




\subsection{Overall Performance (RQ1)}

In this subsection, we provide a comprehensive offline comparison of GIST with various baseline methods, as shown in Table~\ref{tbl:offline}. We summarize our observations as follows: First, incorporating cross-domain lifelong behavior sequences significantly enhances model performance compared to single-domain models (e.g., DIN), which leads to an AUC improvement of at least $0.273\%$. Second, using soft search for searching outperforms hard search, indicating that a more precise GSU module has a distinct advantage in capturing strongly correlated items, which is beneficial for CTR prediction. Finally, by combining lifelong behavioral sequences with joint representation for searching and adding similarity information, our proposed GIST outperforms all baselines, confirming its effectiveness in enhancing cross-domain CTR prediction.

\begin{table}[]
\centering
\caption{Ablation of encoders in the Content-Behavior Joint Training Module. We report AUC in this table.}
\label{tbl:encoder}
\centering
\begin{tabular}{c|cc}
\toprule[2pt]
Method        & AUC & AUC gain \\ \midrule 
Content-only  & 0.7706 & -      \\
Behavior-only & 0.7698 & -0.104\%      \\
Joint         & 0.7720 & 0.181\%     \\
\bottomrule[2pt]
\end{tabular}
\end{table}

\begin{table}[]
\centering
\caption{
Ablation of encoders in the Content-Behavior Joint Training Module. We report Recall@K in this table.}
\label{tbl:encoder_recall}
\centering
\begin{tabular}{c|ccc}
\toprule[2pt]
Method        & Recall@10 & Recall@100 & Recall@1000 \\ \midrule 
Content-only  & 0.1901 & 0.4607 & 0.7470      \\ 
Behavior-only & 0.1189 & 0.2210 & 0.3181   \\ 
Joint         & 0.2112 & 0.5211 & 0.8129    \\
\bottomrule[2pt]
\end{tabular}
\end{table}

\subsection{Module Analyses (RQ2)}

In this subsection, we conduct comprehensive ablation studies to
demonstrate the effectiveness of our key designs, including the Content-Behavior Joint Training module (CBJT) and the Asymmetric Similarity Integration strategy (ASI). 


    
    
    


\subsubsection{\textbf{Analyses of the CBJT Module}}
For the Content-Behavior Joint Training Module, we design ablation experiments to to demonstrate the impact of content- and behavior-based encoders and union training module.


\textbf{Ablation of Encoders.} During experiments, we keep the other components consistent while only making changes to the encoders in the CBJT module. For instance, we only keep the content-based encoder in \textit{Content-only}. Considering that a single encoder does not allow for the fusion, we employ an MLP to learn the behavioral signals from the ESU pairs instead.






The results are shown in Table~\ref{tbl:encoder} and Table~\ref{tbl:encoder_recall}. 
 From these results, we observe that: (1) In Table~\ref{tbl:encoder_recall}, the Recall of the behavior-based encoder (\textit{Behavior-only}) is significantly lower than that of the content encoder, demonstrating that representations derived solely from the behavior-based encoder have weak retrieval capabilities. In contrast, our content-behavior joint representations yield the best results. This underscores that while behavioral information may not perform well in independent retrieval, its effective fusion with content data can greatly enhance vector retrieval capabilities. (2) the Recall metric shares similar trends in performance with the AUC metric. Given that calculating AUC in ranking models demands substantial computational resources and time, we choose Recall as the evaluation metric for subsequent experiments.

\textbf{Ablation of Union Train Module.} 
As discussed above, merely using a behavior encoder does not yield representations with strong retrieval capabilities. Therefore, we design the Union Training Module (UTM) to fusion content-based and behavior-based embeddings. We conduct a series of ablation experiments on UTM, with results presented in the Table~\ref{tbl:utm}. From this table, we observe that without the assistance of the gate network, the model's Recall@1000 decreases by $0.0739$; similarly, without the cross network, Recall@1000 drops by $0.0907$. This indicates the significant importance of both the gate network and the cross network within the Union Training Module for effectively fusing content-behavior representations.

\begin{table}[]
\centering
\caption{Comparison of learning signals.}
\label{tbl:esu}
\centering
\begin{tabular}{c|cc}
\toprule[2pt]
Method        & AUC & AUC gain \\ \midrule 
Swing Pairs  & 0.7711 & -      \\
ESU Pairs & 0.7720 & 0.117\%      \\
\bottomrule[2pt]
\end{tabular}
\end{table}

\textbf{Comparison of Learning Signals.}
In our CBJT, we use ESU pairs as learning signals to 
fuse representations encoded by behavior and content. In previous works, item-to-item pairs based on collaborative filtering, such as swing pairs, were typically extracted as learning signals. Therefore, we conducted a comparison by replacing the learning signals in the Union Training Module with Swing pairs obtained through co-occurrence. The results, as shown in Table~\ref{tbl:esu}, indicate that representations obtained using ESU pairs yield an improvement of 0.117\% AUC over those using swing pairs. This suggests that, compared to the statistically-based swing pairs, ESU pairs can more accurately extract personalized signals, enhancing the model's ability to capture user interests.

\begin{table}[]
\caption{Ablation of the Union Training Module (UTM).}
\label{tbl:utm}
\centering
\small
\begin{tabular}{cccccc}
\toprule[2pt]
\multicolumn{2}{c}{ Components of UTM} & & \multicolumn{3}{c}{ Metrics } \\
\cline { 1 - 2 } \cline { 4 - 6 } Gate & Cross & & Recall@10 & Recall@100 & Recall@1000 \\
\hline$\times$ & $\checkmark$ & & 0.1736 & 0.4417 & 0.7390 \\
$\checkmark$ & $\times$ & & 0.1751 & 0.4401 & 0.7285 \\
$\checkmark$ & $\checkmark$ & & 0.2112 & 0.5211 & 0.8129 \\
\bottomrule[2pt]
\end{tabular}
\end{table}

\begin{figure}[t]
  \centering
  \includegraphics[width=\linewidth]{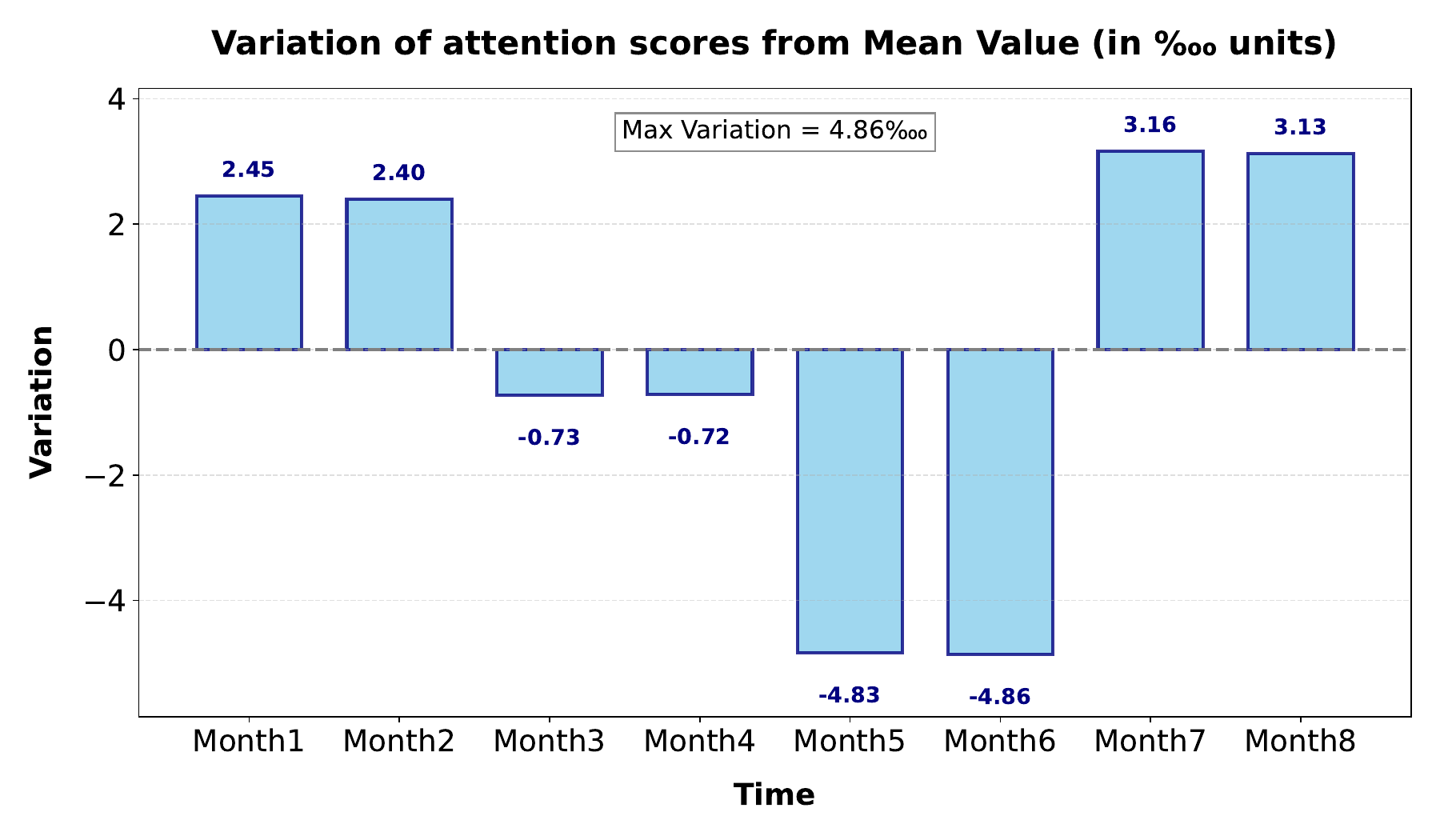}
  \caption{The variation of attention distribution over time.}
  \label{fig:score_dis}
\end{figure}

\begin{figure}[t]
    \centering

    \includegraphics[width=0.8\linewidth]{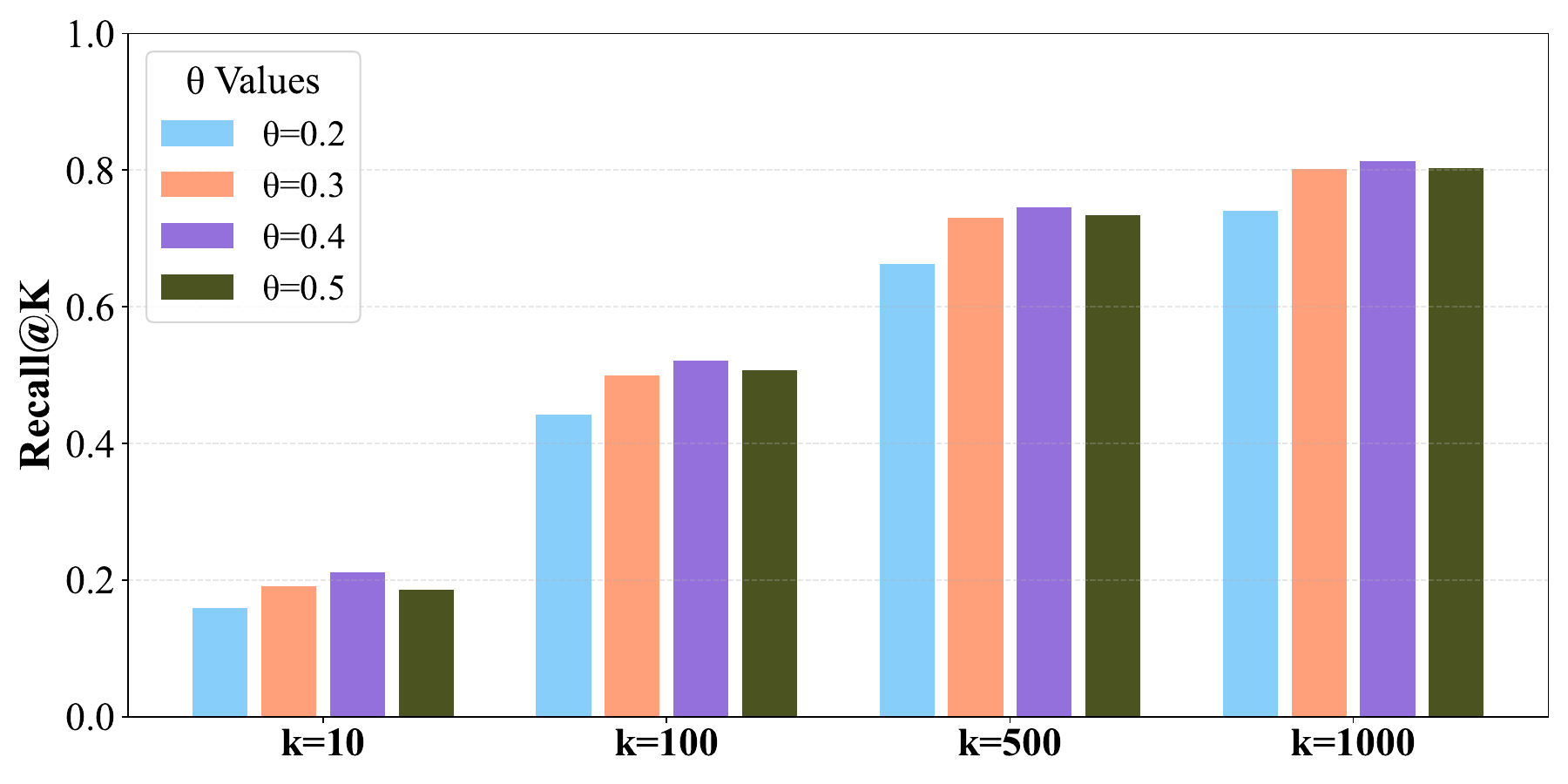}
    
    \caption{We choose ESU pairs based on different threshold $\theta$ and subsequently conduct a Recall@K
    evaluation.}
    \label{fig:recall}
\end{figure}

\textbf{Hyperparameters.} We conducted experimental analysis on the selection threshold $\theta$ for ESU i2i pairs within the same i2i dataset, with results depicted in Figure~\ref{fig:recall}. From this, we derived two key findings. Firstly, at $\theta=0.5$, the model underperformed compared to $\theta=0.4$. This suggests that a higher threshold reduces the number of effective pairs, thereby making it difficult for the CBJT module to learn and optimize effectively. Secondly, at $\theta=0.2$ and $\theta=0.3$, the model's performance again lagged behind that at $\theta=0.4$. This indicates that a lower threshold introduces significant noise, incorporating unconfident pairs into the training data, which negatively impacts the model’s training performance and positions it in a suboptimal state. In summary, $\theta=0.4$ is considered the optimal selection threshold in our experiments. To validate the consistent effectiveness of a fixed threshold $\theta$ across different time periods, we conducted experimental analysis on the distribution of attention scores over consecutive months. Specifically, we represent the distribution of attention scores by the mean value of $a_{k}$ (denoted in Equation~\ref{equ:ak}, the maximum attention score for each sample) from all samples collected on that particular month. As shown in Figure~\ref{fig:score_dis}, after several months of iteration, the mean of the maximum attention scores remains within a very stable distribution, with the maximum variation not exceeding 0.05\%. This supports the rationale for selecting a fixed $\theta$.


\begin{table}[]
\centering
\caption{Ablation of the Asymmetric Similarity Integration strategy (ASI).}
\label{tbl:simi}
\centering
\begin{tabular}{c|cc}
\toprule[2pt]
Method        & AUC & AUC gain \\ \midrule 
Embedding  & 0.7709 & -     \\
SimScore & 0.7718 & 0.117\%      \\
SimScore+SimDist         & 0.7720 & 0.142\%     \\
\bottomrule[2pt]
\end{tabular}
\end{table}

\subsubsection{\textbf{Analyses of the ASI Module}}

We also compare our proposed Asymmetric Similarity Integration strategy (ASI) with other integration methods, which primarily rely on vectors or similarity scores. The overall results, as shown in Table~\ref{tbl:simi}, lead to the following conclusions: Firstly, the lighter similarity information outperforms the use of origin joint embeddings, with experimental results showing an AUC improvement of $0.117\%$. Secondly, by introducing similarity distribution on the target item side, the model gains a more comprehensive perception of behavioral sequences, which also contributes to performance enhancement. These results confirm the effectiveness of our proposed ASI in integrating joint representations into ID-based models.

\subsection{Performance on Items with Different Popularity (RQ3)}

To evaluate the generalization capability of content-behavior joint representations across items with different popularity, we assessed the relative improvements among item groups categorized by their popularity levels. Specifically, we labeled items with user interactions below a certain threshold as the low-interaction group, while those above the threshold were classified as the high-interaction group. We compared GIST's performance with the previous SIM Soft (attention) model. The results shown in Table~\ref{tbl:group} indicate that the joint representation exhibits significant improvements across all groups, demonstrating the effectiveness of GIST on both popular and long-tail items. Notably, we observed even more pronounced enhancements in performance within the low-interaction group, which includes long-tail and cold-start items. These results support our motivation that adaptive integration of multi-modal content signals and behavioral signals is essential for performance gain across cold-start
and interaction-rich items.

\begin{table}[]
\centering
\caption{Performance gain in different item groups.}
\label{tbl:group}
\centering
\begin{tabular}{c|ccc}
\toprule[2pt]
Method        & low-interaction & high-interaction & overall \\ \midrule 
GIST & 0.282\% & 0.196\% & 0.247\%      \\
\bottomrule[2pt]
\end{tabular}
\end{table}

\subsection{Online A/B Test (RQ4)}

In this section, we further test GIST to justify its effectiveness in
real-world CTR prediction systems. Specifically, we conduct a 7-day online A/B test in the advertising scenario of Xiaohongshu (RedNote), which serves the major traffic of hundreds of millions of daily active users. The overall results are significant, GIST achieves a $3.1406\%$ increase in CTR, a $2.1505\%$ growth in Income, and a $1.2528\%$ increase in Cost Per Mille (CPM). In the context of our advertising scenario, even a $0.4\%$
improvement in income is considered statistically significant and represents a meaningful enhancement in performance. These substantial gains clearly demonstrate the effectiveness and superiority of GIST in addressing the issue of sparse behavior in the advertising domain.

\section{Related Work}
\textbf{Cross-domain recommendation (CDR)} involves techniques designed to enhance performance in a low-resource target domain by leveraging knowledge from source domains~\cite{hu2018conet, ma2019pi, zhang2022keep, hou2024cross}. For instance, CoNet~\cite{hu2018conet} facilitates dual knowledge transfer by establishing cross-mapping connections between hidden layers of two base networks, thereby enhancing the flow of cross-domain information. In sequence modeling, CDR techniques have been integrated. Pi-Net~\cite{ma2019pi} and PSJNet~\cite{sun2021parallel} incorporate gating mechanisms to transmit information among shared users. Recent developments~\cite{hou2024cross} have concentrated particularly on the modeling of lifelong sequences in the cross-domain. Some approaches also incorporate graph neural networks~\cite{liu2020cross, li2023preference}. However, most of these strategies focus on short sequences and use a joint training paradigm. A notable advancement in \textbf{sequence modeling} is SIM~\cite{pi2020search}, which breaks down the process into General Search Units (GSU) and Exact Search Units (ESU), thus managing lifelong sequences with reduced computational costs. Subsequent research has expanded upon this framework, achieving significant improvements ~\cite{cao2022sampling, chang2023twin, chen2021end, qin2020user}. Nonetheless, these methods do not completely overcome the unique challenges of cross-domain lifelong sequence modeling, as performance can still suffer due to sparse data in the target domain.

\section{Conclusion}
In this paper, we introduce GIST, a guided content-behavior distillation framework for cross-domain Click-Through Rate (CTR) prediction. By fusing content and behavior signals through the Content-Behavior Joint Training module, GIST learns joint representations that enable effective knowledge transfer across different domains. Subsequently, GIST uses the joint representations to search a fraction of items that are most relevant to user interests, producing cross-domain lifelong sequences. Moreover, we propose an Asymmetric Similarity Integration strategy, which integrates similarity scores and distributions between the target item and users' historical interaction items to augment knowledge transfer. Extensive offline and online experiments demonstrate
the effectiveness of GIST in real-world industrial settings.

\bibliographystyle{ACM-Reference-Format}
\bibliography{union}


\begin{thebibliography}{46}


\ifx \showCODEN    \undefined \def \showCODEN     #1{\unskip}     \fi
\ifx \showISBNx    \undefined \def \showISBNx     #1{\unskip}     \fi
\ifx \showISBNxiii \undefined \def \showISBNxiii  #1{\unskip}     \fi
\ifx \showISSN     \undefined \def \showISSN      #1{\unskip}     \fi
\ifx \showLCCN     \undefined \def \showLCCN      #1{\unskip}     \fi
\ifx \shownote     \undefined \def \shownote      #1{#1}          \fi
\ifx \showarticletitle \undefined \def \showarticletitle #1{#1}   \fi
\ifx \showURL      \undefined \def \showURL       {\relax}        \fi
\providecommand\bibfield[2]{#2}
\providecommand\bibinfo[2]{#2}
\providecommand\natexlab[1]{#1}
\providecommand\showeprint[2][]{arXiv:#2}

\bibitem[Cao et~al\mbox{.}(2022a)]%
        {cao2022disencdr}
\bibfield{author}{\bibinfo{person}{Jiangxia Cao}, \bibinfo{person}{Xixun Lin}, \bibinfo{person}{Xin Cong}, \bibinfo{person}{Jing Ya}, \bibinfo{person}{Tingwen Liu}, {and} \bibinfo{person}{Bin Wang}.} \bibinfo{year}{2022}\natexlab{a}.
\newblock \showarticletitle{Disencdr: Learning disentangled representations for cross-domain recommendation}. In \bibinfo{booktitle}{\emph{Proceedings of the 45th International ACM SIGIR conference on research and development in information retrieval}}. \bibinfo{pages}{267--277}.
\newblock


\bibitem[Cao et~al\mbox{.}(2022b)]%
        {cao2022sampling}
\bibfield{author}{\bibinfo{person}{Yue Cao}, \bibinfo{person}{Xiaojiang Zhou}, \bibinfo{person}{Jiaqi Feng}, \bibinfo{person}{Peihao Huang}, \bibinfo{person}{Yao Xiao}, \bibinfo{person}{Dayao Chen}, {and} \bibinfo{person}{Sheng Chen}.} \bibinfo{year}{2022}\natexlab{b}.
\newblock \showarticletitle{Sampling is all you need on modeling long-term user behaviors for CTR prediction}. In \bibinfo{booktitle}{\emph{Proceedings of the 31st ACM International Conference on Information \& Knowledge Management}}. \bibinfo{pages}{2974--2983}.
\newblock


\bibitem[Chang et~al\mbox{.}(2023)]%
        {chang2023twin}
\bibfield{author}{\bibinfo{person}{Jianxin Chang}, \bibinfo{person}{Chenbin Zhang}, \bibinfo{person}{Zhiyi Fu}, \bibinfo{person}{Xiaoxue Zang}, \bibinfo{person}{Lin Guan}, \bibinfo{person}{Jing Lu}, \bibinfo{person}{Yiqun Hui}, \bibinfo{person}{Dewei Leng}, \bibinfo{person}{Yanan Niu}, \bibinfo{person}{Yang Song}, {et~al\mbox{.}}} \bibinfo{year}{2023}\natexlab{}.
\newblock \showarticletitle{TWIN: TWo-stage interest network for lifelong user behavior modeling in CTR prediction at kuaishou}. In \bibinfo{booktitle}{\emph{Proceedings of the 29th ACM SIGKDD Conference on Knowledge Discovery and Data Mining}}. \bibinfo{pages}{3785--3794}.
\newblock


\bibitem[Chen et~al\mbox{.}(2021b)]%
        {chen2021user}
\bibfield{author}{\bibinfo{person}{Lei Chen}, \bibinfo{person}{Fajie Yuan}, \bibinfo{person}{Jiaxi Yang}, \bibinfo{person}{Xiangnan He}, \bibinfo{person}{Chengming Li}, {and} \bibinfo{person}{Min Yang}.} \bibinfo{year}{2021}\natexlab{b}.
\newblock \showarticletitle{User-specific adaptive fine-tuning for cross-domain recommendations}.
\newblock \bibinfo{journal}{\emph{IEEE Transactions on Knowledge and Data Engineering}} \bibinfo{volume}{35}, \bibinfo{number}{3} (\bibinfo{year}{2021}), \bibinfo{pages}{3239--3252}.
\newblock


\bibitem[Chen et~al\mbox{.}(2021a)]%
        {chen2021end}
\bibfield{author}{\bibinfo{person}{Qiwei Chen}, \bibinfo{person}{Changhua Pei}, \bibinfo{person}{Shanshan Lv}, \bibinfo{person}{Chao Li}, \bibinfo{person}{Junfeng Ge}, {and} \bibinfo{person}{Wenwu Ou}.} \bibinfo{year}{2021}\natexlab{a}.
\newblock \showarticletitle{End-to-end user behavior retrieval in click-through rateprediction model}.
\newblock \bibinfo{journal}{\emph{arXiv preprint arXiv:2108.04468}} (\bibinfo{year}{2021}).
\newblock


\bibitem[Chen et~al\mbox{.}(2024)]%
        {chen2024survey}
\bibfield{author}{\bibinfo{person}{Shu Chen}, \bibinfo{person}{Zitao Xu}, \bibinfo{person}{Weike Pan}, \bibinfo{person}{Qiang Yang}, {and} \bibinfo{person}{Zhong Ming}.} \bibinfo{year}{2024}\natexlab{}.
\newblock \showarticletitle{A Survey on Cross-Domain Sequential Recommendation}.
\newblock \bibinfo{journal}{\emph{arXiv preprint arXiv:2401.04971}} (\bibinfo{year}{2024}).
\newblock


\bibitem[Chen et~al\mbox{.}(2020)]%
        {chen2020simple}
\bibfield{author}{\bibinfo{person}{Ting Chen}, \bibinfo{person}{Simon Kornblith}, \bibinfo{person}{Mohammad Norouzi}, {and} \bibinfo{person}{Geoffrey Hinton}.} \bibinfo{year}{2020}\natexlab{}.
\newblock \showarticletitle{A simple framework for contrastive learning of visual representations}. In \bibinfo{booktitle}{\emph{International conference on machine learning}}. PMLR, \bibinfo{pages}{1597--1607}.
\newblock


\bibitem[Chen and He(2021)]%
        {chen2021exploring}
\bibfield{author}{\bibinfo{person}{Xinlei Chen} {and} \bibinfo{person}{Kaiming He}.} \bibinfo{year}{2021}\natexlab{}.
\newblock \showarticletitle{Exploring simple siamese representation learning}. In \bibinfo{booktitle}{\emph{Proceedings of the IEEE/CVF conference on computer vision and pattern recognition}}. \bibinfo{pages}{15750--15758}.
\newblock


\bibitem[Deng et~al\mbox{.}(2024)]%
        {deng2024end}
\bibfield{author}{\bibinfo{person}{Xiuqi Deng}, \bibinfo{person}{Lu Xu}, \bibinfo{person}{Xiyao Li}, \bibinfo{person}{Jinkai Yu}, \bibinfo{person}{Erpeng Xue}, \bibinfo{person}{Zhongyuan Wang}, \bibinfo{person}{Di Zhang}, \bibinfo{person}{Zhaojie Liu}, \bibinfo{person}{Guorui Zhou}, \bibinfo{person}{Yang Song}, {et~al\mbox{.}}} \bibinfo{year}{2024}\natexlab{}.
\newblock \showarticletitle{End-to-end training of Multimodal Model and ranking Model}.
\newblock \bibinfo{journal}{\emph{arXiv preprint arXiv:2404.06078}} (\bibinfo{year}{2024}).
\newblock


\bibitem[Feng et~al\mbox{.}(2024)]%
        {feng2024long}
\bibfield{author}{\bibinfo{person}{Ningya Feng}, \bibinfo{person}{Junwei Pan}, \bibinfo{person}{Jialong Wu}, \bibinfo{person}{Baixu Chen}, \bibinfo{person}{Ximei Wang}, \bibinfo{person}{Qian Li}, \bibinfo{person}{Xian Hu}, \bibinfo{person}{Jie Jiang}, {and} \bibinfo{person}{Mingsheng Long}.} \bibinfo{year}{2024}\natexlab{}.
\newblock \showarticletitle{Long-Sequence Recommendation Models Need Decoupled Embeddings}.
\newblock \bibinfo{journal}{\emph{arXiv preprint arXiv:2410.02604}} (\bibinfo{year}{2024}).
\newblock


\bibitem[Guo et~al\mbox{.}(2017)]%
        {guo2017deepfm}
\bibfield{author}{\bibinfo{person}{Huifeng Guo}, \bibinfo{person}{Ruiming Tang}, \bibinfo{person}{Yunming Ye}, \bibinfo{person}{Zhenguo Li}, {and} \bibinfo{person}{Xiuqiang He}.} \bibinfo{year}{2017}\natexlab{}.
\newblock \showarticletitle{DeepFM: a factorization-machine based neural network for CTR prediction}.
\newblock \bibinfo{journal}{\emph{arXiv preprint arXiv:1703.04247}} (\bibinfo{year}{2017}).
\newblock


\bibitem[Hou et~al\mbox{.}(2024)]%
        {hou2024cross}
\bibfield{author}{\bibinfo{person}{Ruijie Hou}, \bibinfo{person}{Zhaoyang Yang}, \bibinfo{person}{Yu Ming}, \bibinfo{person}{Hongyu Lu}, \bibinfo{person}{Zhuobin Zheng}, \bibinfo{person}{Yu Chen}, \bibinfo{person}{Qinsong Zeng}, {and} \bibinfo{person}{Ming Chen}.} \bibinfo{year}{2024}\natexlab{}.
\newblock \showarticletitle{Cross-Domain LifeLong Sequential Modeling for Online Click-Through Rate Prediction}. In \bibinfo{booktitle}{\emph{Proceedings of the 30th ACM SIGKDD Conference on Knowledge Discovery and Data Mining}}. \bibinfo{pages}{5116--5125}.
\newblock


\bibitem[Hu et~al\mbox{.}(2018)]%
        {hu2018conet}
\bibfield{author}{\bibinfo{person}{Guangneng Hu}, \bibinfo{person}{Yu Zhang}, {and} \bibinfo{person}{Qiang Yang}.} \bibinfo{year}{2018}\natexlab{}.
\newblock \showarticletitle{Conet: Collaborative cross networks for cross-domain recommendation}. In \bibinfo{booktitle}{\emph{Proceedings of the 27th ACM international conference on information and knowledge management}}. \bibinfo{pages}{667--676}.
\newblock


\bibitem[Li et~al\mbox{.}(2021)]%
        {li2021attentive}
\bibfield{author}{\bibinfo{person}{Dongfang Li}, \bibinfo{person}{Baotian Hu}, \bibinfo{person}{Qingcai Chen}, \bibinfo{person}{Xiao Wang}, \bibinfo{person}{Quanchang Qi}, \bibinfo{person}{Liubin Wang}, {and} \bibinfo{person}{Haishan Liu}.} \bibinfo{year}{2021}\natexlab{}.
\newblock \showarticletitle{Attentive capsule network for click-through rate and conversion rate prediction in online advertising}.
\newblock \bibinfo{journal}{\emph{Knowledge-based systems}}  \bibinfo{volume}{211} (\bibinfo{year}{2021}), \bibinfo{pages}{106522}.
\newblock


\bibitem[Li et~al\mbox{.}(2024)]%
        {li2024lite}
\bibfield{author}{\bibinfo{person}{Haoran Li}, \bibinfo{person}{Junqi Liu}, \bibinfo{person}{Zexian Wang}, \bibinfo{person}{Shiyuan Luo}, \bibinfo{person}{Xiaowei Jia}, {and} \bibinfo{person}{Huaxiu Yao}.} \bibinfo{year}{2024}\natexlab{}.
\newblock \showarticletitle{LITE: Modeling Environmental Ecosystems with Multimodal Large Language Models}.
\newblock \bibinfo{journal}{\emph{arXiv preprint arXiv:2404.01165}} (\bibinfo{year}{2024}).
\newblock


\bibitem[Li et~al\mbox{.}(2025)]%
        {li2025hope}
\bibfield{author}{\bibinfo{person}{Haoran Li}, \bibinfo{person}{Yingjie Qin}, \bibinfo{person}{Baoyuan Ou}, \bibinfo{person}{Lai Xu}, {and} \bibinfo{person}{Ruiwen Xu}.} \bibinfo{year}{2025}\natexlab{}.
\newblock \showarticletitle{HoPE: Hybrid of Position Embedding for Length Generalization in Vision-Language Models}.
\newblock \bibinfo{journal}{\emph{arXiv preprint arXiv:2505.20444}} (\bibinfo{year}{2025}).
\newblock


\bibitem[Li et~al\mbox{.}(2023b)]%
        {li2023blip}
\bibfield{author}{\bibinfo{person}{Junnan Li}, \bibinfo{person}{Dongxu Li}, \bibinfo{person}{Silvio Savarese}, {and} \bibinfo{person}{Steven Hoi}.} \bibinfo{year}{2023}\natexlab{b}.
\newblock \showarticletitle{Blip-2: Bootstrapping language-image pre-training with frozen image encoders and large language models}. In \bibinfo{booktitle}{\emph{International conference on machine learning}}. PMLR, \bibinfo{pages}{19730--19742}.
\newblock


\bibitem[Li et~al\mbox{.}(2022)]%
        {li2022blip}
\bibfield{author}{\bibinfo{person}{Junnan Li}, \bibinfo{person}{Dongxu Li}, \bibinfo{person}{Caiming Xiong}, {and} \bibinfo{person}{Steven Hoi}.} \bibinfo{year}{2022}\natexlab{}.
\newblock \showarticletitle{Blip: Bootstrapping language-image pre-training for unified vision-language understanding and generation}. In \bibinfo{booktitle}{\emph{International conference on machine learning}}. PMLR, \bibinfo{pages}{12888--12900}.
\newblock


\bibitem[Li and Tuzhilin(2020)]%
        {li2020ddtcdr}
\bibfield{author}{\bibinfo{person}{Pan Li} {and} \bibinfo{person}{Alexander Tuzhilin}.} \bibinfo{year}{2020}\natexlab{}.
\newblock \showarticletitle{Ddtcdr: Deep dual transfer cross domain recommendation}. In \bibinfo{booktitle}{\emph{Proceedings of the 13th International Conference on Web Search and Data Mining}}. \bibinfo{pages}{331--339}.
\newblock


\bibitem[Li et~al\mbox{.}(2023a)]%
        {li2023preference}
\bibfield{author}{\bibinfo{person}{Yakun Li}, \bibinfo{person}{Lei Hou}, {and} \bibinfo{person}{Juanzi Li}.} \bibinfo{year}{2023}\natexlab{a}.
\newblock \showarticletitle{Preference-aware graph attention networks for cross-domain recommendations with collaborative knowledge graph}.
\newblock \bibinfo{journal}{\emph{ACM Transactions on Information Systems}} \bibinfo{volume}{41}, \bibinfo{number}{3} (\bibinfo{year}{2023}), \bibinfo{pages}{1--26}.
\newblock


\bibitem[Liu et~al\mbox{.}(2024a)]%
        {liu2024mcrpl}
\bibfield{author}{\bibinfo{person}{Hao Liu}, \bibinfo{person}{Lei Guo}, \bibinfo{person}{Lei Zhu}, \bibinfo{person}{Yongqiang Jiang}, \bibinfo{person}{Min Gao}, {and} \bibinfo{person}{Hongzhi Yin}.} \bibinfo{year}{2024}\natexlab{a}.
\newblock \showarticletitle{MCRPL: A Pretrain, Prompt, and Fine-tune Paradigm for Non-overlapping Many-to-one Cross-domain Recommendation}.
\newblock \bibinfo{journal}{\emph{ACM Transactions on Information Systems}} \bibinfo{volume}{42}, \bibinfo{number}{4} (\bibinfo{year}{2024}), \bibinfo{pages}{1--24}.
\newblock


\bibitem[Liu et~al\mbox{.}(2020)]%
        {liu2020cross}
\bibfield{author}{\bibinfo{person}{Meng Liu}, \bibinfo{person}{Jianjun Li}, \bibinfo{person}{Guohui Li}, {and} \bibinfo{person}{Peng Pan}.} \bibinfo{year}{2020}\natexlab{}.
\newblock \showarticletitle{Cross domain recommendation via bi-directional transfer graph collaborative filtering networks}. In \bibinfo{booktitle}{\emph{Proceedings of the 29th ACM international conference on information \& knowledge management}}. \bibinfo{pages}{885--894}.
\newblock


\bibitem[Liu et~al\mbox{.}(2024b)]%
        {liu2024at4ctr}
\bibfield{author}{\bibinfo{person}{Qi Liu}, \bibinfo{person}{Xuyang Hou}, \bibinfo{person}{Defu Lian}, \bibinfo{person}{Zhe Wang}, \bibinfo{person}{Haoran Jin}, \bibinfo{person}{Jia Cheng}, {and} \bibinfo{person}{Jun Lei}.} \bibinfo{year}{2024}\natexlab{b}.
\newblock \showarticletitle{AT4CTR: Auxiliary Match Tasks for Enhancing Click-Through Rate Prediction}. In \bibinfo{booktitle}{\emph{Proceedings of the AAAI Conference on Artificial Intelligence}}, Vol.~\bibinfo{volume}{38}. \bibinfo{pages}{8787--8795}.
\newblock


\bibitem[Ma et~al\mbox{.}(2018)]%
        {ma2018modeling}
\bibfield{author}{\bibinfo{person}{Jiaqi Ma}, \bibinfo{person}{Zhe Zhao}, \bibinfo{person}{Xinyang Yi}, \bibinfo{person}{Jilin Chen}, \bibinfo{person}{Lichan Hong}, {and} \bibinfo{person}{Ed~H Chi}.} \bibinfo{year}{2018}\natexlab{}.
\newblock \showarticletitle{Modeling task relationships in multi-task learning with multi-gate mixture-of-experts}. In \bibinfo{booktitle}{\emph{Proceedings of the 24th ACM SIGKDD international conference on knowledge discovery \& data mining}}. \bibinfo{pages}{1930--1939}.
\newblock


\bibitem[Ma et~al\mbox{.}(2019)]%
        {ma2019pi}
\bibfield{author}{\bibinfo{person}{Muyang Ma}, \bibinfo{person}{Pengjie Ren}, \bibinfo{person}{Yujie Lin}, \bibinfo{person}{Zhumin Chen}, \bibinfo{person}{Jun Ma}, {and} \bibinfo{person}{Maarten~de Rijke}.} \bibinfo{year}{2019}\natexlab{}.
\newblock \showarticletitle{$\pi$-net: A parallel information-sharing network for shared-account cross-domain sequential recommendations}. In \bibinfo{booktitle}{\emph{Proceedings of the 42nd international ACM SIGIR conference on research and development in information retrieval}}. \bibinfo{pages}{685--694}.
\newblock


\bibitem[Oord et~al\mbox{.}(2018)]%
        {oord2018representation}
\bibfield{author}{\bibinfo{person}{Aaron van~den Oord}, \bibinfo{person}{Yazhe Li}, {and} \bibinfo{person}{Oriol Vinyals}.} \bibinfo{year}{2018}\natexlab{}.
\newblock \showarticletitle{Representation learning with contrastive predictive coding}.
\newblock \bibinfo{journal}{\emph{arXiv preprint arXiv:1807.03748}} (\bibinfo{year}{2018}).
\newblock


\bibitem[Ouyang et~al\mbox{.}(2021)]%
        {ouyang2021learning}
\bibfield{author}{\bibinfo{person}{Wentao Ouyang}, \bibinfo{person}{Xiuwu Zhang}, \bibinfo{person}{Shukui Ren}, \bibinfo{person}{Li Li}, \bibinfo{person}{Kun Zhang}, \bibinfo{person}{Jinmei Luo}, \bibinfo{person}{Zhaojie Liu}, {and} \bibinfo{person}{Yanlong Du}.} \bibinfo{year}{2021}\natexlab{}.
\newblock \showarticletitle{Learning graph meta embeddings for cold-start ads in click-through rate prediction}. In \bibinfo{booktitle}{\emph{Proceedings of the 44th International ACM SIGIR Conference on Research and Development in Information Retrieval}}. \bibinfo{pages}{1157--1166}.
\newblock


\bibitem[Ouyang et~al\mbox{.}(2020)]%
        {ouyang2020minet}
\bibfield{author}{\bibinfo{person}{Wentao Ouyang}, \bibinfo{person}{Xiuwu Zhang}, \bibinfo{person}{Lei Zhao}, \bibinfo{person}{Jinmei Luo}, \bibinfo{person}{Yu Zhang}, \bibinfo{person}{Heng Zou}, \bibinfo{person}{Zhaojie Liu}, {and} \bibinfo{person}{Yanlong Du}.} \bibinfo{year}{2020}\natexlab{}.
\newblock \showarticletitle{Minet: Mixed interest network for cross-domain click-through rate prediction}. In \bibinfo{booktitle}{\emph{Proceedings of the 29th ACM international conference on information \& knowledge management}}. \bibinfo{pages}{2669--2676}.
\newblock


\bibitem[Pi et~al\mbox{.}(2020)]%
        {pi2020search}
\bibfield{author}{\bibinfo{person}{Qi Pi}, \bibinfo{person}{Guorui Zhou}, \bibinfo{person}{Yujing Zhang}, \bibinfo{person}{Zhe Wang}, \bibinfo{person}{Lejian Ren}, \bibinfo{person}{Ying Fan}, \bibinfo{person}{Xiaoqiang Zhu}, {and} \bibinfo{person}{Kun Gai}.} \bibinfo{year}{2020}\natexlab{}.
\newblock \showarticletitle{Search-based user interest modeling with lifelong sequential behavior data for click-through rate prediction}. In \bibinfo{booktitle}{\emph{Proceedings of the 29th ACM International Conference on Information \& Knowledge Management}}. \bibinfo{pages}{2685--2692}.
\newblock


\bibitem[Qin et~al\mbox{.}(2020)]%
        {qin2020user}
\bibfield{author}{\bibinfo{person}{Jiarui Qin}, \bibinfo{person}{Weinan Zhang}, \bibinfo{person}{Xin Wu}, \bibinfo{person}{Jiarui Jin}, \bibinfo{person}{Yuchen Fang}, {and} \bibinfo{person}{Yong Yu}.} \bibinfo{year}{2020}\natexlab{}.
\newblock \showarticletitle{User behavior retrieval for click-through rate prediction}. In \bibinfo{booktitle}{\emph{Proceedings of the 43rd International ACM SIGIR Conference on Research and Development in Information Retrieval}}. \bibinfo{pages}{2347--2356}.
\newblock


\bibitem[Radford et~al\mbox{.}(2021)]%
        {radford2021learning}
\bibfield{author}{\bibinfo{person}{Alec Radford}, \bibinfo{person}{Jong~Wook Kim}, \bibinfo{person}{Chris Hallacy}, \bibinfo{person}{Aditya Ramesh}, \bibinfo{person}{Gabriel Goh}, \bibinfo{person}{Sandhini Agarwal}, \bibinfo{person}{Girish Sastry}, \bibinfo{person}{Amanda Askell}, \bibinfo{person}{Pamela Mishkin}, \bibinfo{person}{Jack Clark}, {et~al\mbox{.}}} \bibinfo{year}{2021}\natexlab{}.
\newblock \showarticletitle{Learning transferable visual models from natural language supervision}. In \bibinfo{booktitle}{\emph{International conference on machine learning}}. PMLR, \bibinfo{pages}{8748--8763}.
\newblock


\bibitem[Sheng et~al\mbox{.}(2024)]%
        {sheng2024enhancing}
\bibfield{author}{\bibinfo{person}{Xiang-Rong Sheng}, \bibinfo{person}{Feifan Yang}, \bibinfo{person}{Litong Gong}, \bibinfo{person}{Biao Wang}, \bibinfo{person}{Zhangming Chan}, \bibinfo{person}{Yujing Zhang}, \bibinfo{person}{Yueyao Cheng}, \bibinfo{person}{Yong-Nan Zhu}, \bibinfo{person}{Tiezheng Ge}, \bibinfo{person}{Han Zhu}, {et~al\mbox{.}}} \bibinfo{year}{2024}\natexlab{}.
\newblock \showarticletitle{Enhancing Taobao Display Advertising with Multimodal Representations: Challenges, Approaches and Insights}. In \bibinfo{booktitle}{\emph{Proceedings of the 33rd ACM International Conference on Information and Knowledge Management}}. \bibinfo{pages}{4858--4865}.
\newblock


\bibitem[Sun et~al\mbox{.}(2021)]%
        {sun2021parallel}
\bibfield{author}{\bibinfo{person}{Wenchao Sun}, \bibinfo{person}{Muyang Ma}, \bibinfo{person}{Pengjie Ren}, \bibinfo{person}{Yujie Lin}, \bibinfo{person}{Zhumin Chen}, \bibinfo{person}{Zhaochun Ren}, \bibinfo{person}{Jun Ma}, {and} \bibinfo{person}{Maarten De~Rijke}.} \bibinfo{year}{2021}\natexlab{}.
\newblock \showarticletitle{Parallel split-join networks for shared account cross-domain sequential recommendations}.
\newblock \bibinfo{journal}{\emph{IEEE Transactions on Knowledge and Data Engineering}} \bibinfo{volume}{35}, \bibinfo{number}{4} (\bibinfo{year}{2021}), \bibinfo{pages}{4106--4123}.
\newblock


\bibitem[Swietojanski et~al\mbox{.}(2016)]%
        {swietojanski2016learning}
\bibfield{author}{\bibinfo{person}{Pawel Swietojanski}, \bibinfo{person}{Jinyu Li}, {and} \bibinfo{person}{Steve Renals}.} \bibinfo{year}{2016}\natexlab{}.
\newblock \showarticletitle{Learning hidden unit contributions for unsupervised acoustic model adaptation}.
\newblock \bibinfo{journal}{\emph{IEEE/ACM Transactions on Audio, Speech, and Language Processing}} \bibinfo{volume}{24}, \bibinfo{number}{8} (\bibinfo{year}{2016}), \bibinfo{pages}{1450--1463}.
\newblock


\bibitem[Wang et~al\mbox{.}(2017)]%
        {wang2017deep}
\bibfield{author}{\bibinfo{person}{Ruoxi Wang}, \bibinfo{person}{Bin Fu}, \bibinfo{person}{Gang Fu}, {and} \bibinfo{person}{Mingliang Wang}.} \bibinfo{year}{2017}\natexlab{}.
\newblock \showarticletitle{Deep \& cross network for ad click predictions}.
\newblock In \bibinfo{booktitle}{\emph{Proceedings of the ADKDD'17}}. \bibinfo{pages}{1--7}.
\newblock


\bibitem[Wang et~al\mbox{.}(2021)]%
        {wang2021dcn}
\bibfield{author}{\bibinfo{person}{Ruoxi Wang}, \bibinfo{person}{Rakesh Shivanna}, \bibinfo{person}{Derek Cheng}, \bibinfo{person}{Sagar Jain}, \bibinfo{person}{Dong Lin}, \bibinfo{person}{Lichan Hong}, {and} \bibinfo{person}{Ed Chi}.} \bibinfo{year}{2021}\natexlab{}.
\newblock \showarticletitle{Dcn v2: Improved deep \& cross network and practical lessons for web-scale learning to rank systems}. In \bibinfo{booktitle}{\emph{Proceedings of the web conference 2021}}. \bibinfo{pages}{1785--1797}.
\newblock


\bibitem[Yang et~al\mbox{.}(2021)]%
        {yang2021taco}
\bibfield{author}{\bibinfo{person}{Jianwei Yang}, \bibinfo{person}{Yonatan Bisk}, {and} \bibinfo{person}{Jianfeng Gao}.} \bibinfo{year}{2021}\natexlab{}.
\newblock \showarticletitle{Taco: Token-aware cascade contrastive learning for video-text alignment}. In \bibinfo{booktitle}{\emph{Proceedings of the IEEE/CVF international conference on computer vision}}. \bibinfo{pages}{11562--11572}.
\newblock


\bibitem[Yang et~al\mbox{.}(2022)]%
        {yang2022vision}
\bibfield{author}{\bibinfo{person}{Jinyu Yang}, \bibinfo{person}{Jiali Duan}, \bibinfo{person}{Son Tran}, \bibinfo{person}{Yi Xu}, \bibinfo{person}{Sampath Chanda}, \bibinfo{person}{Liqun Chen}, \bibinfo{person}{Belinda Zeng}, \bibinfo{person}{Trishul Chilimbi}, {and} \bibinfo{person}{Junzhou Huang}.} \bibinfo{year}{2022}\natexlab{}.
\newblock \showarticletitle{Vision-language pre-training with triple contrastive learning}. In \bibinfo{booktitle}{\emph{Proceedings of the IEEE/CVF Conference on Computer Vision and Pattern Recognition}}. \bibinfo{pages}{15671--15680}.
\newblock


\bibitem[Yu et~al\mbox{.}(2020)]%
        {yu2020gradient}
\bibfield{author}{\bibinfo{person}{Tianhe Yu}, \bibinfo{person}{Saurabh Kumar}, \bibinfo{person}{Abhishek Gupta}, \bibinfo{person}{Sergey Levine}, \bibinfo{person}{Karol Hausman}, {and} \bibinfo{person}{Chelsea Finn}.} \bibinfo{year}{2020}\natexlab{}.
\newblock \showarticletitle{Gradient surgery for multi-task learning}.
\newblock \bibinfo{journal}{\emph{Advances in Neural Information Processing Systems}}  \bibinfo{volume}{33} (\bibinfo{year}{2020}), \bibinfo{pages}{5824--5836}.
\newblock


\bibitem[Yuan et~al\mbox{.}(2023)]%
        {yuan2023go}
\bibfield{author}{\bibinfo{person}{Zheng Yuan}, \bibinfo{person}{Fajie Yuan}, \bibinfo{person}{Yu Song}, \bibinfo{person}{Youhua Li}, \bibinfo{person}{Junchen Fu}, \bibinfo{person}{Fei Yang}, \bibinfo{person}{Yunzhu Pan}, {and} \bibinfo{person}{Yongxin Ni}.} \bibinfo{year}{2023}\natexlab{}.
\newblock \showarticletitle{Where to go next for recommender systems? id-vs. modality-based recommender models revisited}. In \bibinfo{booktitle}{\emph{Proceedings of the 46th International ACM SIGIR Conference on Research and Development in Information Retrieval}}. \bibinfo{pages}{2639--2649}.
\newblock


\bibitem[Zadeh et~al\mbox{.}(2017)]%
        {zadeh2017tensor}
\bibfield{author}{\bibinfo{person}{Amir Zadeh}, \bibinfo{person}{Minghai Chen}, \bibinfo{person}{Soujanya Poria}, \bibinfo{person}{Erik Cambria}, {and} \bibinfo{person}{Louis-Philippe Morency}.} \bibinfo{year}{2017}\natexlab{}.
\newblock \showarticletitle{Tensor fusion network for multimodal sentiment analysis}.
\newblock \bibinfo{journal}{\emph{arXiv preprint arXiv:1707.07250}} (\bibinfo{year}{2017}).
\newblock


\bibitem[Zhang et~al\mbox{.}(2019)]%
        {zhang2019deep}
\bibfield{author}{\bibinfo{person}{Shuai Zhang}, \bibinfo{person}{Lina Yao}, \bibinfo{person}{Aixin Sun}, {and} \bibinfo{person}{Yi Tay}.} \bibinfo{year}{2019}\natexlab{}.
\newblock \showarticletitle{Deep learning based recommender system: A survey and new perspectives}.
\newblock \bibinfo{journal}{\emph{ACM computing surveys (CSUR)}} \bibinfo{volume}{52}, \bibinfo{number}{1} (\bibinfo{year}{2019}), \bibinfo{pages}{1--38}.
\newblock


\bibitem[Zhang et~al\mbox{.}(2022)]%
        {zhang2022keep}
\bibfield{author}{\bibinfo{person}{Yujing Zhang}, \bibinfo{person}{Zhangming Chan}, \bibinfo{person}{Shuhao Xu}, \bibinfo{person}{Weijie Bian}, \bibinfo{person}{Shuguang Han}, \bibinfo{person}{Hongbo Deng}, {and} \bibinfo{person}{Bo Zheng}.} \bibinfo{year}{2022}\natexlab{}.
\newblock \showarticletitle{KEEP: An industrial pre-training framework for online recommendation via knowledge extraction and plugging}. In \bibinfo{booktitle}{\emph{Proceedings of the 31st ACM International Conference on Information \& Knowledge Management}}. \bibinfo{pages}{3684--3693}.
\newblock


\bibitem[Zhao et~al\mbox{.}(2023)]%
        {zhao2023cross}
\bibfield{author}{\bibinfo{person}{Chuang Zhao}, \bibinfo{person}{Hongke Zhao}, \bibinfo{person}{Ming He}, \bibinfo{person}{Jian Zhang}, {and} \bibinfo{person}{Jianping Fan}.} \bibinfo{year}{2023}\natexlab{}.
\newblock \showarticletitle{Cross-domain recommendation via user interest alignment}. In \bibinfo{booktitle}{\emph{Proceedings of the ACM Web Conference 2023}}. \bibinfo{pages}{887--896}.
\newblock


\bibitem[Zhou et~al\mbox{.}(2018)]%
        {zhou2018deep}
\bibfield{author}{\bibinfo{person}{Guorui Zhou}, \bibinfo{person}{Xiaoqiang Zhu}, \bibinfo{person}{Chenru Song}, \bibinfo{person}{Ying Fan}, \bibinfo{person}{Han Zhu}, \bibinfo{person}{Xiao Ma}, \bibinfo{person}{Yanghui Yan}, \bibinfo{person}{Junqi Jin}, \bibinfo{person}{Han Li}, {and} \bibinfo{person}{Kun Gai}.} \bibinfo{year}{2018}\natexlab{}.
\newblock \showarticletitle{Deep interest network for click-through rate prediction}. In \bibinfo{booktitle}{\emph{Proceedings of the 24th ACM SIGKDD international conference on knowledge discovery \& data mining}}. \bibinfo{pages}{1059--1068}.
\newblock


\bibitem[Zhu et~al\mbox{.}(2022)]%
        {zhu2022personalized}
\bibfield{author}{\bibinfo{person}{Yongchun Zhu}, \bibinfo{person}{Zhenwei Tang}, \bibinfo{person}{Yudan Liu}, \bibinfo{person}{Fuzhen Zhuang}, \bibinfo{person}{Ruobing Xie}, \bibinfo{person}{Xu Zhang}, \bibinfo{person}{Leyu Lin}, {and} \bibinfo{person}{Qing He}.} \bibinfo{year}{2022}\natexlab{}.
\newblock \showarticletitle{Personalized transfer of user preferences for cross-domain recommendation}. In \bibinfo{booktitle}{\emph{Proceedings of the fifteenth ACM international conference on web search and data mining}}. \bibinfo{pages}{1507--1515}.
\newblock


\end{thebibliography}


\end{document}